\documentclass[10pt, a4paper]{article}

\usepackage[final]{lrec2026} 

\title{Navigating Global AI Regulation: A Multi-Jurisdictional Retrieval-Augmented Generation System\thanks{Preprint. Accepted at PoliticalNLP Workshop, LREC 2026}}

\name{Courtney Ford\textsuperscript{1}, Ojas Rane\textsuperscript{2,3}, Susan Leavy\textsuperscript{1,2}} 

\address{\textsuperscript{1}School of Information \& Communication Studies, University College Dublin, Ireland,\\ 
\textsuperscript{2}Insight Centre for Data Analytics, Dublin, Ireland, \\
\textsuperscript{3}School of Bimolecular \& Biomedical Science, University College Dublin, Ireland, \\
         \{courtney.ford, ojas.rane, susan.leavy\}@ucd.ie\\}

\abstract{
Navigating AI regulation across jurisdictions is increasingly difficult for policymakers, legal professionals, and researchers. To address this, we present a multi-jurisdictional Retrieval-Augmented Generation system for global AI regulation. Our corpus includes 242 documents across 68 jurisdictions, ranging from formal legislation like the EU AI Act to unstructured policy documents such as national AI strategies. The system makes three technical contributions: type-specific chunking that preserve legal structure across heterogenous documents; conditional retrieval routing with entity detection and metadata for legal citations; and priority-based re-ranking to boost enacted legislation over policy and secondary sources. Evaluation of 50 queries reveals strong performance across both single-entity and multi-jurisdictional questions, achieving 0.87 average faithfulness and 0.84 average answer relevancy. Single-entity queries achieve 0.86 average faithfulness and 0.92 average answer relevancy, while multi-jurisdictional comparison queries achieve 0.88 average faithfulness and 0.75 average answer relevancy. These findings highlight the effectiveness of domain-specific retrieval strategies for navigating complex, heterogenous regulatory corpora. 
 \\ \newline  }

\begin{document}

\maketitleabstract

\section{Introduction}
Advances in AI capabilities have intensified longstanding concerns about AI safety~\cite{bengio2026international, weidinger2021ethicalsocialrisksharm, weidinger2022}, privacy~\cite{gupta2023, Yang_2026}, and societal impact~\cite{Kido_Takadama_2024, Liu2024}. Many governments have shifted from suggested AI ethical development guidelines toward binding regulation. The EU AI Act~\cite{EUAIAct2024}, enacted in 2024, represents a landmark development in this shift, and regulatory frameworks are emerging across major jurisdictions worldwide.

Navigating this landscape is increasingly difficult. Stakeholders including policymakers, legal professionals, researchers, and civil society organisations, face challenges due to fragmentation across jurisdictions~\cite{zaidan2024ai}, volume~\cite{oecd2025governing}, and divergent jurisdictional approaches~\cite{Alanoca_2025} to AI governance. Understanding how different jurisdictions approach the same issue requires reading across multiple lengthy, technical documents, often published in their native languages, to identify similarities and differences. Existing repositories such as the OECD AI Policy Observatory~\cite{oecdai} provide valuable cataloguing and high-level summaries, but do not support interactive querying of specific documents or comparison of how multiple jurisdictions address particular regulatory issues. 

We present a multi-jurisdictional Retrieval-Augmented Generation (RAG) system to address this gap. By "multi-jurisdictional," we refer to the system's ability to handle regulatory documents from multiple countries and international bodies, enabling both single-jurisdiction queries and comparative analysis across jurisdictions. To address the need for interactive, detailed access to regulatory texts that static repositories cannot provide, our system enables users to ask questions about individual articles and compare how different jurisdictions address similar regulatory issues. The system makes three key contributions: first, type-specific chunking strategies that preserve legal structure for formal legislation while appropriately handling unstructured policy documents; second, a conditional retrieval pipeline that uses entity detection, direct metadata lookup for legal citations, and automatic fallback to EU-level regulations when member state documents are unavailable; and third, tailored re-ranking strategies that prioritise primary legislation for single-entity queries while ensuring balanced representations across jurisdictions for comparative queries. We evaluate the system on 50 queries spanning article lookups, single-jurisdiction questions, and multiple jurisdictional comparisons using the Retrieval Augmented Generation Assessment (RAGAs) framework~\cite{es-etal-2024-ragas}. The system achieves 0.87 average faithfulness and 0.84 average answer relevancy, with single-entity queries performing strongly (0.86 faithfulness, 0.92 answer relevancy) while multi-jurisdictional queries reveal interesting challenges (0.88 faithfulness, 0.75 answer relevancy). 

\section{Related Work}
Retrieval-Augmented Generation (RAG) addresses the limitations of parametric knowledge in pre-trained language models by incorporating external knowledge sources at inference time~\cite{lewis2020retrieval}. This approach has been increasingly applied to legal and regulatory domains, where specialised, up-to-date information is critical for tasks ranging from regulatory compliance to legal question answering.

\subsection{RAG for Legal and Regulatory Documents}
Legal and regulatory domains present unique challenges for RAG systems due to the complex, hierarchical structure of documents and the need for precise citation~\cite{hindi2025}. Within AI policy and regulatory contexts specifically, ~\citet{kalra-etal-2024-hypa} introduce HyPA-RAG, a hybrid parameter-adaptive system designed for AI legal applications. Testing on New York City Local Law 144, they demonstrate that adaptive retrieval strategies combining dense, sparse, and knowledge graph methods significantly improve response fidelity and contextual precision for regulatory queries. Addressing broader regulatory compliance challenges, ~\citet{kim2024} developed a QA-RAG system for pharmaceutical guidelines that outperforms conventional RAG methods on Food and Drug Administration (FDA) and International Council for Harmonisation (ICH) regulatory documents through hybrid retrieval using both original queries and hypothetical answers. ~\citet{gokhan2024riragregulatoryinformationretrieval} target financial regulatory compliance, creating the ObliQA dataset of 27,869 question-passage pairs from Abu Dhabi Global Markets documents and proposing RePASs, an evaluation metric designed to assess whether generated answers accurately capture regulatory obligations. Beyond regulatory contexts, ~\citet{zhang2025} demonstrate that iterative query refinement improves recall in legal contract analysis, achieving 78.67\% recall through multi-round RAG compared to 74.67\% with single-round approaches. 

Recent work has developed specialised evaluation frameworks and benchmarks for legal RAG systems. ~\citet{li2025lexrag} introduce LexRAG, the first benchmark specifically designed for multi-turn legal consultations, providing 1,013 dialogue samples across 17,228 legal articles to evaluate both conversational knowledge retrieval and response generation. ~\citet{pipitone2024legalbench} develop LegalBench-RAG to assess retrieval precision in legal RAG systems, emphasising the extraction of minimal, highly relevant text segments to avoid exceeding context windows and reducing hallucinations. ~\citet{keisha2025} propose a comprehensive evaluation framework combining RAGAs, BERTScore, and ROUGE metrics to assess both semantic alignment and faithfulness in legal RAG outputs. 

A key technical challenge in legal RAG is preserving document structure during chunking and indexing. Legal documents are hierarchically organised into articles and provisions with cross-references, and standard character-based chunking methods risk breaking these structural units. Recent work addresses this through hierarchy-aware approaches: ~\citet{ferraris2024legal} find that standard chunking methods achieve low semantic closeness scores for legal retrieval, ~\citet{reuter2025towards} propose summary-augmented chunking to preserve global context, and ~\citet{akarajaradwong-etal-2025-nitibench} treat complete legal sections as single chunks to maintain legal concepts. 

While these systems demonstrate the effectiveness of RAG for legal and regulatory tasks, our work addresses a distinct challenge: enabling interactive retrieval and comparison across AI regulatory documents from multiple jurisdictions. This requires handling heterogenous document types and supporting comparative queries that synthesise information across different regulatory frameworks.

\section{Methodology}

\subsection{Corpus and Data Collection}
Our corpus consists of 242 regulatory documents spanning 68 jurisdictions, including enacted and proposed legislation, and policy documents such as national AI strategies and governance frameworks. While we aimed to balance geographic representation, coverage is constrained by the availability of formalised AI regulatory documents. As shown in Table~\ref{tab:corpus_distribution}, certain regions have more publicly accessible enacted legislation and formal policy frameworks. Of the 68 jurisdictions, 45 are represented by a single document, which is expected given that most countries require only one or two primary regulatory documents while supranational entities such as the EU are represented by multiple documents spanning distinct regulatory domains. This distribution reflects differences in regulatory approaches and document availability rather than regulatory activity levels, and directly impacts system performance on cross-regional comparisons (see Section 5.2). 

The corpus is predominantly in English, with some documents (n=47) in their original language. Each document is tagged with metadata including jurisdiction (entity), title, publication year, language, status (e.g., enacted, draft, proposed), and document type.

\begin{table}[!ht]
\begin{center}
\begin{tabularx}{\columnwidth}{|X|c|c|}
\hline
\textbf{Region} & \textbf{Jurisdictions} & \textbf{Documents} \\
\hline
North America & 4 & 95 \\
\hline
Asia & 17 & 67 \\
\hline
Europe & 23 & 48 \\
\hline
Africa & 11 & 13 \\
\hline
International & 7 & 10 \\
\hline
South America & 4 & 6 \\
\hline
Oceania & 2 & 3 \\
\hline
\textbf{Total} & \textbf{68} & \textbf{242} \\
\hline
\end{tabularx}
\caption{Corpus distribution by region.}
\label{tab:corpus_distribution}
\end{center}
\end{table}

\subsection{Document Pre-Processing and Chunking}
A challenge in building multi-jurisdictional legal corpora is structural heterogeneity. Our 242-document corpus spans formal legislative texts (e.g., EU AI Act, GDPR, US congressional bills) with explicit article or section markers, and unstructured policy documents (e.g., national AI strategies, white papers) written in continuous prose. We implemented two chunking strategies:

\textbf{Structured Documents (n=81)}: For legislative texts, we treat each structural unit (article, section, provision) as a single chunk to avoid splitting provisions across chunks  and enable precise citation. Structural boundaries are identified via document-specific parsers, with metadata preserved (e.g., "Article 5(3)", "Section 1798.100"). When a single unit exceeds 2000 characters, we apply recursive splitting with 100-character overlap.

\textbf{Unstructured Documents (n=160)}: For policy documents and white papers, we apply RecursiveCharacterTextSplitter with 1000-character chunks and 200-character overlap, using hierarchical separators (paragraphs to sentences to words). The larger overlap compensates for absent structural markers by maintaining contextual continuity. 

This total process yielded 38,750 chunks, each retaining metadata for jurisdiction, document title, year, language, and (where applicable) structural references for citation generation. 

\subsection{Embedding and Vectorisation}
We use the all-mpnet-base-v2 model (768-dimensional, 110M parameters) from the sentence-transformers library~\cite{reimers-gurevych-2019-sentence} for document embedding. This model provides strong semantic understanding and handles multilingual content effectively, which is critical for our international corpus. 

All 38,750 chunks are embedded with L2 normalisation and indexed using Faiss~\cite{douze2025faisslibrary} with a flat L2 index. While approximate nearest neighbour methods offer faster retrieval, we prioritise accuracy through exhaustive search given the relatively modest corpus size. The Faiss docstore integrates chunk metadata, enabling filtered retrieval by jurisdiction, document type, or other attributes during query processing. 

\subsection{Retrieval Pipeline}
Multi-jurisdictional queries present unique challenges: users may reference jurisdictions using country names, adjectives, or abbreviations; queries may focus on specific legal provisions or compare multiple jurisdictions; and semantic search can prioritise secondary sources (reports about regulations, e.g., Center for AI and Digital Policy [CAIDP]: Artificial Intelligence and Democratic Values Index [AIDV]~\cite{caidp2025index}) over primary legislation. We address these issues through a multi-stage conditional retrieval pipeline (Figure~\ref{pipeline}). 

\subsubsection{Query Analysis} We first analyse queries to extract structural cues. We use a two-stage approach to identify jurisdictions, with regex-based word-boundary matching against our 68 entity names handling most cases. For queries using adjectival forms ("Chinese regulations") or alternative names, we fall back to GPT-4o-mini with prompt-based entity extraction, constrained to return only entities present in the corpus metadata. This hybrid approach balances speed (regex path for most queries) with coverage (LLM for edge cases), and ensures that entity detection cannot surface jurisdictions outside the corpus.  

\subsubsection{Conditional Routing Strategy} 
Based on query analysis, we route to one of three pathways: 

\textbf{Pathway A: Direct Metadata Lookup:} When a query references a specific article or section within a known document, we bypass semantic search entirely and retrieve chunks directly via metadata filtering on article/section numbers and document names. 

\textbf{Pathway B: Entity-Filtered Semantic Search:} For single-jurisdiction queries, we retrieve $k \times 5 \times 5$ candidates via similarity search (over-retrieving to ensure recall before filtering), filter to the target jurisdiction, then re-rank. For EU member states with no country-specific results, we automatically expand to include EU-level regulations, acknowledging that EU directives apply to member states. 

\textbf{Pathway C: Multi-Jurisdiction Search:} For comparison queries, we disable entity filtering and document-name boosting, retrieving based purely on semantic similarity to allow documents from multiple jurisdictions to be ranked by relevance.

\subsubsection{Re-Ranking Strategies} 
A critical challenge in the development of this system was that semantic search often ranked secondary sources higher than primary legislation. For single-entity queries (Pathways A and B), we address this through multiplicative score adjustment based on document characteristics, where lower scores indicate better matches in Faiss. Enacted legislation receives the strongest boost (score $\times$ 0.6), followed by proposed or introduced legislation (score $\times$ 0.8). Other documents including AI strategies, policy documents, and white papers receive no adjustment, relying on semantic similarity alone to determine their ranking. Documents referenced by name in the user query receive an additional boost (score $\times$ 0.7) before re-ranking. 

For multi-jurisdictional comparison queries (Pathway C), a different re-ranking strategy is applied. Document-name boosting is disabled to prevent entity detection from inadvertently prioritising one jurisdiction's documents over another. Retrieved documents are then grouped by jurisdiction and re-ranked by taking the highest-scoring chunk from each jurisdiction in turn, repeating until $k$ results are filled. This ensures balanced representation across all requested frameworks rather than allowing a single jurisdiction's primary legislation to dominate the retrieved context. 

The pipeline returns the top-$k$ re-ranked results with citation metadata for generation.

\begin{figure}[!ht]
\begin{center}
\includegraphics[width=.90\columnwidth]{pipeline.png}
\caption{Conditional retrieval pipeline. Query analysis routes to one of three pathways: direct metadata lookup for article-specific queries (Path A), entity-filtered semantic search (Path B), or multi-jurisdiction search for comparison queries (Path C). Paths A and B apply priority-based re-ranking to boost primary legislation, while Path C applies comparison-aware re-ranking to ensure balanced representation across queried jurisdictions.}
\label{pipeline}
\end{center}
\end{figure}

\section{Evaluation}
We evaluate the system using the RAGAs framework~\cite{es-etal-2024-ragas}, which provides automatic evaluation metrics for RAG systems. We assess performance on two metrics: \textit{faithfulness}, measuring whether generated answers are grounded in retrieved sources rather than hallucinated, and \textit{answer relevancy}, measuring whether generated answers directly address the question asked. We evaluate on 50 test queries divided into two categories: single-entity queries (n=25) and multi-jurisdictional comparison queries (n=25). Single-entity queries comprise article-specific questions (n=8), conceptual questions about individual jurisdictions (n=10), and EU member state queries that test the automatic fallback mechanism described in Pathway B in Section 3.4.2 (n=7). Multi-jurisdictional queries comprise straightforward comparisons between jurisdictions with strong corpus coverage (n=15) and harder comparisons testing entity detection with adjectival and alternate jurisdiction forms (n=10). The full list of test queries see Appendix A.

\subsection{Faithfulness}
Faithfulness measures the proportion of claims in a generated answer that can be attributed to retrieved source documents. This metric assesses whether the system grounds its responses in evidence rather than hallucinating information. 

For each test query, we generate answers using our retrieval pipeline, including query analysis, conditional routing strategy, priority-based re-ranking, and answer generation via GPT-4o. We extract the top-5 re-ranked source chunks that were actually provided to the language model as context. RAGAs employs GPT-4 to decompose each generated answer into individual claims and verify whether each claim is supported by the retrieved contexts. Scores range from 0 (fully unsupported) to 1 (all claims grounded in sources), with scores above 0.8 indicating strong faithfulness. 

\textbf{Results:} Table ~\ref{tab:faithfulness} shows faithfulness scores across 49 successfully evaluated queries. Our system achieved an overall average faithfulness score of 0.87, indicating that the majority of generated claims are well-supported by retrieved sources. One query could not be evaluated due to failed entity detection at the retrieval stage, resulting in no retrieved context. 

Performance was strong across both query categories, with single-entity queries achieving 0.86 average faithfulness and multi-jurisdictional queries achieving 0.88. This demonstrates that the system maintains strong grounding in source documents regardless of query complexity. 

\begin{table}[!ht]
\begin{center}
\begin{tabularx}{\columnwidth}{|X|c|c|}
\hline
\textbf{Query Category} & \textbf{n} & \textbf{Avg. Faithfulness} \\
\hline
Single-entity & 25 & 0.86 \\
\hline
Multi-jurisdictional & 24 & 0.88 \\
\hline
\textbf{Overall} & \textbf{49} & \textbf{0.87} \\
\hline
\end{tabularx}
\caption{Faithfulness scores by query category. Scores range from 
0 (fully unsupported) to 1 (all claims grounded in sources), with 
scores above 0.8 indicating strong grounding in source documents.}
\label{tab:faithfulness}
\end{center}
\end{table}

\subsection{Answer Relevancy}
Answer relevancy measures whether generated answers directly address the specific question asked. While faithfulness assesses grounding in sources, answer relevancy evaluates whether the response is actually useful to the user. This metric is particularly important for multi-jurisdictional comparison queries, where incomplete retrieval may result in answers that are truthful but fail to address the full scope of the question. 

RAGAs employs GPT-4 to assess answer relevancy by comparing the generated answer against the original question, scoring how well the response addresses what was asked. Scores range from 0 (completely irrelevant) to 1 (fully addresses the question). 

\textbf{Results:} Table ~\ref{tab:relevancy} shows answer relevancy scores across 49 queries. Our system achieved an overall average answer relevancy score of 0.84. Performance varied by query type: single-entity queries achieved 0.92 average answer relevancy, while multi-jurisdictional queries achieved 0.75. 

This difference reflects the challenge of multi-jurisdictional retrieval. When the system successfully retrieves documents from all requested jurisdictions, comparison queries achieve high relevancy. However, when retrieval provides documents from only one jurisdiction, the system reports the limitation (maintaining high faithfulness) but cannot fully address the comparison (resulting in lower relevancy). 

\begin{table}[!ht]
\begin{center}
\begin{tabularx}{\columnwidth}{|X|c|c|}
\hline
\textbf{Query Category} & \textbf{n} & \textbf{Avg. Relevancy} \\
\hline
Single-entity & 25 & 0.92 \\
\hline
Multi-jurisdictional & 24 & 0.75 \\
\hline
\textbf{Overall} & \textbf{49} & \textbf{0.84} \\
\hline
\end{tabularx}
\caption{Answer relevancy scores by query category. Scores range from 
0 (completely irrelevant) to 1 (fully addresses the question).}
\label{tab:relevancy}
\end{center}
\end{table}

\subsection{Query Examples}
\subsubsection{Single-Entity Queries}
\textbf{Query:} "What are Japan's AI governance guidelines?"

The system retrieved relevant documents and generated an answer explaining Japan's "guidelines of guidelines" approach, soft law framework, and seven principles from the Social Principles of Human-centric AI~\cite{japan2019social}. The retrieved chunks contained comprehensive policy descriptions that directly addressed the query. This answer achieved 1.0 faithfulness and 0.97 answer relevancy.
\\\\
\textbf{Query:} "What does Article 17 of the GDPR say?"

The system retrieved fragmented chunks from GDPR~\cite{eu2016gdpr} Article 17 and generated an answer correctly identifying the "Right to erasure" and explaining three key conditions: when data is no longer necessary, when consent is withdrawn, and when the data subject objects to processing. The answer provided accurate legal interpretation despite the retrieved chunks being sentence fragments rather than complete article text. This answer achieved 0.5 faithfulness and 0.96 answer relevancy.

\subsubsection{Multi-Jurisdictional Queries}
\textbf{Query:} "Compare the UK and Canada's approaches to AI regulation."

The system retrieved documents from both jurisdictions and generated a structured comparison contrasting Canada's emerging legislative framework (the Artificial Intelligence and Data Act~\cite{canada2022aida} establishing a Commissioner and emphasising systemic effects monitoring) with the UK's sector-specific approach (relying on existing regulatory bodies to issue guidance). The answer achieved 0.8 faithfulness and 0.97 answer relevancy, with the lower faithfulness score reflecting some synthesised characterisations of regulatory approaches that extended slightly beyond the literal source text.
\\\\
\textbf{Query:} "How do Egyptian and Indian national AI strategies compare?"

The system retrieved documents describing Egypt's National AI Strategy in detail (including its four pillars: AI for Government, AI for Development, Governance, and Data Infrastructure) but found only brief mentions of India's approach without substantive strategic details. The answer stated: "The provided sources do not contain specific information about India's national AI strategy, so a direct comparison cannot be fully made" before describing Egypt's strategy and noting that India has "elaborated national plans with clear targets." This answer achieved 0.86 faithfulness and 0.0 answer relevancy, reflecting the pattern where incomplete retrieval results in incomplete responses.

\section{Discussion}

Our evaluation demonstrates strong overall performance (0.87 faithfulness, 0.84 answer relevancy across 50 queries), but reveals a key finding: single-entity and multi-jurisdictional queries exhibit fundamentally different performance patterns. Understanding this divergence provides insights into the challenges of multi-jurisdictional retrieval-augmented generation. 

\subsection{Faithfulness and Answer Relevancy}
Single-entity queries achieved aligned metrics (0.86 faithfulness, 0.92 answer relevancy), indicating that when the system retrieves documents from the correct jurisdiction, it can both ground its response in those sources and directly address the question asked. The Japan AI governance example (Section 4.3.1) illustrates this: comprehensive policy documents enabled a detailed, accurate answer achieving near-perfect scores on both metrics. 

Multi-jurisdictional queries, however, exhibited divergence (0.88 faithfulness, 0.75 answer relevancy). While faithfulness remained high, answer relevancy dropped substantially. This pattern reflects a fundamental challenge where the system retrieves grounded content from available sources but often cannot retrieve from all requested jurisdictions simultaneously. 

The Egyptian-Indian comparison (Section 4.3.2) illustrates the extreme case of this pattern. The system retrieved Egyptian documents but only briefly mentioned India's approach without substantive details, despite India being represented by two documents in the corpus. This retrieval failure reflects a limitation of semantic similarity scoring rather than corpus sparsity, and suggests that query decomposition into separate single-jurisdiction subqueries may better ensure balanced retrieval in such cases.

Even successful comparisons seem affected by this challenge. The UK-Canada example (Section 4.3.2) achieved high answer relevancy (0.97) but lower faithfulness (0.8), with the faithfulness penalty reflecting synthesised characterisations that extend beyond literal source text. These characterisations are a necessary component of comparative analysis but one that automatic metrics flag as insufficiently grounded. 

These findings reveal a fundamental limitation of semantic similarity for multi-jurisdictional retrieval: semantic search inherently favours depth over breadth, retrieving multiple highly-similar chunks from one jurisdiction rather than distributing retrieval across multiple jurisdictions. Our re-ranking mechanism (Section 3.4.3) addresses this to some extent by ensuring balanced representation among retrieved documents, but cannot solve retrieval failures when documents from one jurisdiction never appear in the initial candidate pool due to lower semantic similarity scores. 

\subsection{Limitations and Future Work}
The system's behaviour reflects a design choice to prioritise accuracy over completeness. When documents from only one jurisdiction are retrieved, the system reports this limitation rather than fabricating information. In legal and regulatory contexts, incomplete truth is preferable to fabrication, however, this results in incomplete answers to users' comparative questions. Answer relevancy's penalty for honest incompleteness may not align with the ethical requirements of legal RAG systems, suggesting that comparison queries may require different evaluation criteria than single-entity queries.

Our reliance on automatic evaluation metrics also introduces limitations. The GDPR Article 17 example (Section 4.3.1) achieved high answer relevancy (0.96) by correctly explaining the "Right to erasure," but received low faithfulness (0.5) because retrieved chunks were sentence fragments rather than complete text. The automatic metric penalised accurate legal interpretation for not matching the source word-for-word, highlighting that RAGAs measures grounding but not legal correctness or real-world utility. We are currently conducting user evaluation with legal researchers and policy analysts to address this gap. 

The entity detection component successfully identified jurisdictions in 49 of 50 evaluation queries, with one query failing at the entity detection stage and returning no retrieved context. Zero answer relevancy scores observed in five further queries reflect downstream retrieval failures rather than entity detection errors. In these cases, jurisdictions were correctly identified but retrieved documents lacked sufficient coverage of the specific topic queried. 

The uneven corpus distribution (Table~\ref{tab:corpus_distribution}) further limits system performance on cross-regional comparisons. Jurisdictions with sparse document coverage are less likely to be successfully retrieved in multi-jurisdictional queries, as demonstrated by failed comparison queries involving underrepresented African and South American jurisdictions. Expanding coverage in these regions depends on the availability of formalised regulatory documents, which varies significantly across jurisdictions.

Future work should address these limitations through several directions. Query decomposition, which would involve breaking multi-jurisdictional queries into separate single-jurisdiction sub-queries and then synthesising, may better ensure balanced representation of multiple entities. Formal evaluation of the entity detection component, including precision and recall across the diversity of legal naming conventions and adjectival forms in our corpus, would also strengthen confidence in the retrieval pipeline, as detection errors can propagate through subsequent stages. Alternative retrieval strategies specifically designed for comparative queries may also be necessary to achieve the breadth that comparison queries require. 

\section{Conclusion} 
We present a multi-jurisdictional RAG system for AI regulatory documents spanning 68 jurisdictions, designed to enable policymakers, legal professionals, and researchers to interactively query and compare regulatory frameworks across borders. The system makes three contributions. First, type-specific chunking preserves the hierarchical structure of formal legislation while handling unstructured policy documents, ensuring legal provisions are not split across chunks. Second, a conditional retrieval pipeline routes queries to the appropriate retrieval strategy based on whether they target a specific article, a single jurisdiction, or multiple jurisdictions for comparison. Third, tailored re-ranking ensures primary legislation is prioritised for single-entity queries while guaranteeing balanced representation across jurisdictions for comparative analysis. Evaluation on 50 queries demonstrates strong performance across both single-entity and multi-jurisdictional queries. Performance was nevertheless variable, with retrieval challenges, including corpus coverage gaps and semantic similarity limitations, the primary source of lower scores across both query types. 

\section{Acknowledgments}
This publication has emanated from research funded by the European Union under grant number (101107969). Views and opinions expressed are however those of the author(s) only and do not necessarily reflect those of the European Union or Erasmus+. Neither
the European Union nor the granting authority can be held responsible for them. The research was also supported by Research Ireland under grant number (12/RC/2289\_P2). For the purpose of Open Access, the author has applied a CC BY public copyright licence to any Author Accepted Manuscript version arising from this submission. 

\section*{Appendix A. Evaluation Queries}

\subsection*{Single-Entity Queries (n=25)}
\subsubsection*{Article-Specific Queries (n=8)}
\begin{enumerate}
    \item What does Article 5 of the EU AI Act say?
    \item What does Article 30 of the EU AI Act say?
    \item What does Article 52 of the EU AI Act say?
    \item What does Article 9 of the EU AI Act say?
    \item What does Article 13 of the EU AI Act say?
    \item What does Article 16 of the GDPR say?
    \item What does Article 25 of the GDPR say?
    \item What does Article 17 of the GDPR say?
\end{enumerate}

\subsubsection*{Conceptual Single-Jurisdiction Queries (n=10)}
\begin{enumerate}
    \item What are high-risk AI systems according to the EU AI Act?
    \item What is Singapore's Model AI Governance Framework?
    \item What are Japan's AI governance guidelines?
    \item How does China regulate algorithmic recommendations?
    \item What is Canada's approach to automated decision-making?
    \item What are Australia's AI ethics principles?
    \item What is Brazil's AI strategy?
    \item What is Egypt's national AI strategy?
    \item What are the G7 Hiroshima Process principles for AI?
    \item What is India's national AI strategy?
\end{enumerate}

\subsubsection*{EU Member State Queries (n=7)}
\begin{enumerate}
    \item What is Germany's AI regulation?
    \item How does France regulate AI?
    \item What are Ireland's AI governance rules?
    \item What is Denmark's national AI strategy?
    \item What is Estonia's approach to AI governance?
    \item What are Austria's AI ethics guidelines?
    \item What is Czechia's national AI strategy?
\end{enumerate}

\subsection*{Multi-Jurisdictional Queries (n=25)}
\subsubsection*{Straightforward Comparisons (n=15)}
\begin{enumerate}
    \item How does the EU approach to AI regulation differ from the US?
    \item Compare China's AI regulation to Japan's approach.
    \item How do transparency requirements differ between the EU AI 
    Act and Singapore's framework?
    \item Compare the UK and Canada's approaches to AI regulation.
    \item How do Australia and India approach AI ethics differently?
    \item Compare Singapore and South Korea's AI governance frameworks.
    \item How do Canada and Japan approach automated decision-making 
    differently?
    \item Compare Brazil and India's national AI strategies.
    \item How do the US and UK approach AI governance differently?
    \item Compare Australia and Singapore's AI governance principles.
    \item How do China and South Korea approach AI regulation 
    differently?
    \item Compare the G7 Hiroshima principles and the EU AI Act.
    \item How do Canada and Australia approach AI ethics differently?
    \item Compare India and Singapore's national AI strategies.
    \item How do the UK and Japan approach AI governance differently?
\end{enumerate}

\subsubsection*{Harder Comparisons with Adjectival and Alternate 
Forms (n=10)}
\begin{enumerate}
    \item How do Chinese AI regulations compare to European ones?
    \item What are the differences between American and European 
    approaches to AI governance?
    \item Compare Japanese and Korean AI governance frameworks.
    \item How do Singaporean and Australian AI governance principles 
    differ?
    \item What is the British approach to AI governance compared 
    to Canada?
    \item How do Canadian and American approaches to AI regulation 
    differ?
    \item Compare Indian and Brazilian national AI strategies.
    \item How do African Union and Asian approaches to AI governance 
    differ?~\footnote{This query failed at the entity detection stage and 
returned no retrieved context, and is therefore excluded from 
the RAGAs evaluation results reported in Section 4.}
    \item Compare the G7 and EU approaches to AI governance.
    \item How do Egyptian and Indian national AI strategies compare?
\end{enumerate}

\section{Bibliographical References}\label{sec:reference}

\bibliographystyle{lrec2026-natbib}
\bibliography{final_sub}


\end{document}